\documentclass[pdflatex,sn-mathphys-num]{sn-jnl}

\usepackage{graphicx}%
\usepackage{multirow}%
\usepackage{amsmath,amssymb,amsfonts}%
\usepackage{amsthm}%
\usepackage{mathrsfs}%
\usepackage[title]{appendix}%
\usepackage{xcolor}%
\usepackage{textcomp}%
\usepackage{manyfoot}%
\usepackage{booktabs}%
\usepackage{algorithm}%
\usepackage{algorithmicx}%
\usepackage{algpseudocode}%
\usepackage{listings}%
\usepackage{rotating}    
\usepackage{float}  
\usepackage{soul}  

\definecolor{yellowhl}{RGB}{255,255,128} 

\begin{document}

\title[Article Title]{Investigating potential causes of Sepsis with Bayesian network structure learning}


\author*[1]{\fnm{Bruno} \sur{Petrungaro}}\email{b.petrungaro@qmul.ac.uk}

\author[1]{\fnm{Neville K.} \sur{Kitson}}\email{n.k.kitson@qmul.ac.uk}

\author[1]{\fnm{Anthony C.} \sur{Constantinou}}\email{a.constantinou@qmul.ac.uk}

\affil*[1]{\orgdiv{Bayesian Artificial Intelligence research lab, MInDS research group}, \orgname{Queen Mary University of London}, \orgaddress{\street{Mile End Road}, \city{London}, \postcode{E1 4NS}, \country{UK}}}

\abstract{\emph{Sepsis} is a life-threatening and serious global health issue. This study combines knowledge with available hospital data to investigate the potential causes of \emph{Sepsis} that can be affected by policy decisions. We investigate the underlying causal structure of this problem by combining clinical expertise with score-based, constraint-based, and hybrid structure learning algorithms. A novel approach to model averaging and knowledge-based constraints was implemented to arrive at a consensus structure for causal inference. The structure learning process highlighted the importance of exploring data-driven approaches alongside clinical expertise. This includes discovering unexpected, although reasonable, relationships from a clinical perspective. Hypothetical interventions on \emph{Chronic Obstructive Pulmonary Disease}, \emph{Alcohol dependence}, and \emph{Diabetes} suggest that the presence of any of these risk factors in patients increases the likelihood of \emph{Sepsis}. This finding, alongside measuring the effect of these risk factors on \emph{Sepsis}, has potential policy implications. Recognising the importance of prediction in improving health outcomes related to \emph{Sepsis}, the model is also assessed in its ability to predict \emph{Sepsis} by evaluating accuracy, sensitivity, and specificity. These three indicators all had results around 70\%, and the AUC was 80\%, which means the causal structure of the model is reasonably accurate given that the models were trained on data available for commissioning purposes only.}

\keywords{Directed acyclic graphs, causal discovery, causal inference, model averaging}



\maketitle

\section{Introduction}\label{sec1}

The Third International Consensus for \emph{Sepsis} was held in 2016, where \cite{bib1} define \emph{Sepsis} as a "\textit{life-threatening organ dysfunction caused by a dysregulated host response to infection}". The World Health Organization declared \emph{Sepsis} a global health priority in 2017 (\cite{bib2}). Despite being known since the time of Hippocrates (460-370 BC) in Ancient Greece (\cite{bib2}), \emph{Sepsis} is still a leading cause of mortality due to limited treatment options. \cite{bib3} quantify the level of mortality by stating that 13 million people are diagnosed with \emph{Sepsis} each year worldwide, and 4 million people die because of it. In the United Kingdom, \cite{bib3} report that \emph{Sepsis} causes a figure slightly below 37,000 deaths per year, which is larger than the deaths caused by lung cancer or those caused by breast and bowel cancer combined.

Extensive research has been conducted in building predictive models for the early diagnosis of \emph{Sepsis}. Some examples include \cite{bib4}, \cite{bib5}, \cite{bib6}, \cite{bib7}, \cite{bib8} and \cite{bib9}. These models, which include Bayesian Networks and some constructed using supervised learning models, are intended to aid the management of \emph{Sepsis} in hospitals and prevent severe cases by supporting the early detection of \emph{Sepsis}. Although these models may help to reduce the incidence and mortality of \emph{Sepsis} due to their early prediction capabilities, they do not investigate potential areas for intervention to reduce the incidence of \emph{Sepsis}. Furthermore, all the predictive models reviewed use biomarkers or clinical data for predicting \emph{Sepsis}. However, these data are not routinely collected by the National Health Service (NHS) in England for service commissioning purposes. This lack of data may be true for commissioners of services in other healthcare systems. Accessing patient records without a clinical or direct care reason is generally restricted. Even if commissioners could access these data, they probably would not have the necessary expertise to interpret it if they do not have a clinical background. Because commissioners play a crucial role in service design, building a model using data they regularly access, like the one developed in this study, is important for exploring new services that can improve patient outcomes.

The methods used for prediction in the public health sector are generally regression-based, as described in \cite{bib8} and \cite{bib9}. One problem with regression models is that they do not enable accurate modelling of the often-interdependent relationships of the risk factors for patients developing \emph{Sepsis}. The relationships between these factors might be highly complex, and it is essential to model them as such. 

Regression models do not generally explore possible causal relationships. In contrast, this study examines graphs learnt from data and explores their potential to reveal causal relationships. However, these algorithms rely on assumptions which often do not hold in practice. Here, we attempt to reduce the resulting errors by incorporating expert knowledge and averaging models produced by different algorithms.

Causal Bayesian Networks (CBNs) are represented by a Directed Acyclic Graph (DAG), which provides an intuitive and easy way to read the causal relationships between risk factors and model the impact of interventions available to policymakers. DAGs consist of nodes representing variables and arcs (or edges) assumed to represent causal relationships. The acyclicity condition in DAGs formalises the principle that a variable cannot cause itself. The DAG representation in CBNs provides a clear visualisation of complex clinical relationships and an intuitive framework for identifying causal pathways. The structure learning process described in the previous paragraph determines the structure of these arcs and nodes. One of the relevant features of CBNs is that they enable us to model predictive and diagnostic inference and the effect of hypothetical interventions using the so-called do-calculus framework proposed by \cite{bib10}.

This study makes the following contributions:
\begin{itemize}
    \item Develops a CBN model using data routinely collected by the NHS in England for commissioning purposes.
    \item Captures the highly complex interrelations amongst risk factors with a novel method combining expert knowledge and averaging six structure learning algorithms.
    \item Assesses the ability of the model to predict the presence of \emph{Sepsis}.
    \item Identifies which risk factors can be influenced by policy decisions and uses algorithms to discover and estimate their causal effects on \emph{Sepsis}.
\end{itemize}

Unlike purely predictive methods, the proposed CBN approach reveals how modifiable risk factors can be targeted to reduce the incidence of \textit{Sepsis}. This focus on actionable insights is critical in policymaking contexts, where commissioners require evidence of which interventions can yield the most significant impact on patient outcomes. In addition, we explore averaging structures across multiple data-driven structures to reduce uncertainty and use expert domain knowledge to judge disagreements between algorithms, thereby mitigating the limitations of purely algorithmic solutions. These limitations originate from the algorithmic reliance on assumptions that may not hold in real-world healthcare settings.

This work bridges a critical gap between technical innovation and practical applicability by introducing a causally grounded, policy-focused approach to \textit{Sepsis}. The result is a well-suited framework for healthcare commissioners, equipping them with predictive capabilities and causal insights to inform strategic resource allocation and service design. Although built explicitly for \textit{Sepsis}, we hope the framework used to create this model inspires researchers to apply it to other diseases.
 
\section{Collating data using clinical knowledge}\label{sec2}

\subsection{Factors Associated with Sepsis}

A range of risk factors is identified in the literature as being associated with the presence of \emph{Sepsis}. The risk factors described below can be found in routinely collected data for service commissioning purposes by the NHS. The discussion did not include risk factors that cannot be found in these data. Table 1 lists these factors and studies identifying them as likely to be linked to \emph{Sepsis}. Our study will consider all of these factors.

\begin{table}[h]
\caption{Risk factors of Sepsis as identified by the literature review}\label{tab1}%
\begin{tabular}{@{}ll@{}}
\toprule
\textbf{Risk Factor}&{\textbf{Reference}} \\
\midrule
White blood cell counts (WBC) & \cite{bib8} \\
Glucose & \cite{bib8} \\
Sex & \cite{bib11} \\
Age & \cite{bib12} \\
Urinary catheters (UC) &	\cite{bib12} \\
Central venous lines (CVL) & \cite{bib12} \\
Mechanical ventilation (MV) & \cite{bib12} \\ & \cite{bib13}  \\
Fluid resuscitation (FR) & \cite{bib13} \\
Vasoactive drugs (VD) &	\cite{bib13} \\
Emergency surgeries (ES) & \cite{bib13} \\ & \cite{bib14} \\
Parenteral nutrition (PN) & \cite{bib14} \\
Low serum albumin level at admission (Albumin) &	\cite{bib14} \\
Number of surgical interventions (Number of Procedures) &	\cite{bib14} \\
Abdominal surgery (AS) &	\cite{bib14} \\
Presence of two or more comorbidities & \cite{bib14} \\
Coma & \cite{bib14} \\
Chronic Obstructive Pulmonary Disease (COPD) & \cite{bib15} \\
Immunosuppressive disorders (ID) & \cite{bib15} \\
Alcohol dependence (Alcohol) & \cite{bib16}\\
\botrule
\end{tabular}
\end{table}
\noindent

In addition to the factors presented in Table 1, we will also consider \emph{ethnicity} (also referred to as \emph{ethnic group}) as a factor. Although no substantial evidence was found about its relationship with 
\emph{Sepsis}, \cite{bib17} point out that ethnic minorities suffer worse health outcomes. We will also consider the \emph{number of diagnoses} registered for a patient as a factor. Since the number of surgical interventions (\emph{number of Interventions, Operations, and Procedures} will be used) is included as a risk factor, which can be interpreted as a surrogate measure of the complexity of a patient's case, the natural question arises whether the \emph{number of diagnoses} might also be relevant in this context. Moreover, the \emph{number of diagnoses} also represents the presence of two or more comorbidities since diagnoses include comorbidities. Therefore, the \emph{number of diagnoses} will be included in this study instead of the presence of two or more comorbidities.

Table 2 presents other variables that we incorporated based on clinical expertise in this study. Clinical expertise was elicited through collaboration with Dr Jonathon Dean, a clinical expert based at the North East London Adult Critical Care Operational Delivery Network and the co-founder of the North East London Critical Care Transfer and Retrieval service. Dr Dean has spent the last five years working in anaesthetics and critical care in North East London, where \emph{Sepsis} is an important and topical pathology.

\begin{table}[h]
\caption{Additional risk factors considered in this study based on clinical expertise}\label{tab2}%
\begin{tabular}{@{}l@{}}
\toprule
\textbf{Risk Factor} \\
\midrule
Elevated respiratory rate (ERR) \\	
Hypotension \\
Impaired swallow and aspiration (ISA)	\\
Trauma \\
Reduced oxygen saturation (ROS) \\
Tachycardia \\
Antibiotics	\\
Elevated lactate level (ELL) \\
Myelodysplastic disorders (Myelodysplastic) \\ 
Cystic fibrosis \\
Diabetes \\	
Cancer \\
Anuria \\	
Chronic Kidney Disease Stage 3+ (CKD Stage 3+) \\
\botrule
\end{tabular}
\end{table}
\noindent

\subsection{Mechanisms and Pathophysiology of Sepsis}

Figure 1, which is based on what is presented in \cite{bib3}, provides a visual understanding of the mechanisms and pathophysiology of \emph{Sepsis}. It summarises what influences occurrences of \emph{Sepsis} in patients. We need the presence of an \emph{infectious agent}, \emph{genetic factors} that might predispose patients to or against \emph{Sepsis}, and the patient's \emph{innate immunity}, which might help prevent it. The CBN constructed in this study will be based on the assumption that the risk factors selected are highly correlated with \emph{genetic factors} and are, therefore, a good representation of them. This is a common assumption that we also adopt in this paper, in that \emph{genetic factors} tend to influence the occurrences of long-term conditions (\cite{bib18}). \emph{Infectious agents} (IA) will be included in the model. However, no data can measure \emph{innate immunity}. The only data available is whether or not patients have \emph{immunosuppressive disorders}, and this information is assumed to be helpful in understanding if patients have their immune system compromised and, therefore, are more likely to develop \emph{Sepsis}.

\begin{figure}[h]
\centerline{\includegraphics[width=8cm, keepaspectratio = true]{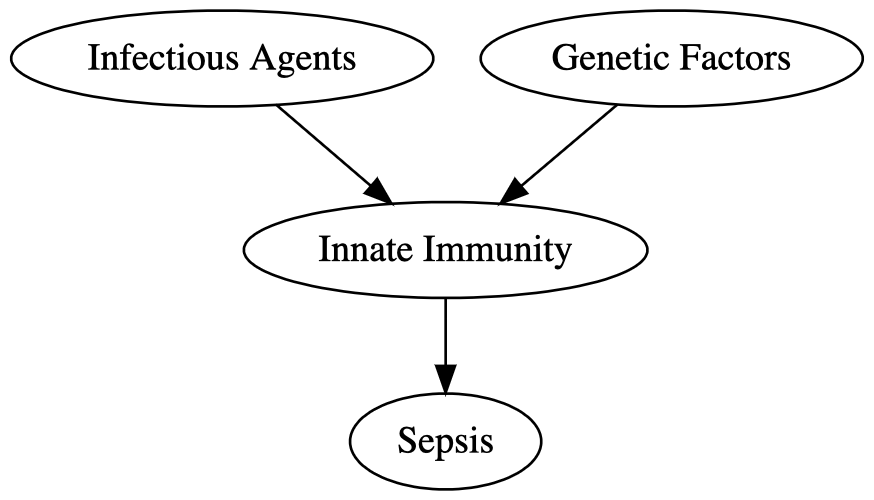}}
\caption{Mechanisms and pathophysiology of Sepsis}
\label{fig}
\end{figure}

\section{Data pre-processing}\label{sec3}

This study will use the Secondary Uses Service (SUS+) database, which includes data for patients admitted to hospitals (inpatients), patients using accident and emergency services, and patients with outpatient appointments within the NHS in England. This study will look at admissions classified as emergencies; this means admissions to hospitals that were not planned. ICD-10 (International Classification of Diseases, 10th Revision)is the 10th version of the global standard for diagnosis classification (\cite{bib19}). These codes are used to classify diagnoses in the database and will be crucial to constructing the variables identified by the literature review and expert knowledge. OPCS codes identify the different interventions, operations, and procedures within the NHS in England, and we have used them to create relevant variables in this study. Variables were constructed with clinical guidance, but only non-clinical authors had access to the database. SUS+ was accessed through the National Commissioning Data Repository (NCDR), a pseudonymised patient-level data repository managed by NHS England.

\subsection{Data Extraction}

The \texttt{odbc} R package (\cite{bib20}) was used to connect to the database. This package also allows for the database to be queried using SQL. Each row in the admitted patients table represents an episode during a hospital spell. A spell can be composed of many episodes. The episodes of a spell are likely to contain similar information regarding diagnosis, procedures and patient information. To obtain only one episode representing any given spell, we applied the rule of selecting the most relevant episode during a spell, where the most relevant is determined by the most relevant Healthcare Resource Groups (HRGs) during the spell. In this way, we ensure we are not double counting the same spells while capturing the most important information about that spell. Treatments are grouped into the same HRGs when they are deemed to require similar levels of resources. Therefore, we selected the episode where the highest level of resources was used for treatment.

The data were extracted for the financial year 2019/2020, which ran from the 1st of April 2019 until the 31st of March 2020. Except for the variables representing the \emph{Age}, \emph{number of diagnoses}, \emph{number of procedures}, \emph{sex}, and \emph{ethnicity} of a patient, all the variables extracted were in the form of a binary indicator establishing the presence or absence of the risk factor. A sample of 1,000,000 episodes was taken due to server constraints; i.e., the runtime for this study on the server already spans multiple days with this sample size.

\subsection{Missing Data}

Structure learning algorithms, including those used in this study, generally require complete datasets. However, the collated data contained missing values, although the amount of missing data is minimal. The proportion of missing values in the variables \emph{age}, \emph{ethnicity}, and \emph{sex}) is less than 0.6\%, i.e. fewer than 6,000 rows of our 1,000,000 sample. 

A common but often problematic solution to missingness is to delete the rows with missing values. This works well under the very strong assumption that the missing data are Missing Completely At Random (MCAR) and if sufficient sample size remains after deletion. We say that data are MCAR when missingness is unrelated to other variables and to observed and unobserved values of the variable itself. A much more likely assumption to make for missing data in this study is that data are Missing At Random (MAR). Data is MAR when the fact that data is missing is related to other variables in the dataset but not to the variable itself. The data we use in this study is a cross-section of hospital episodes. Therefore, it is more reasonable to assume that the missing data are MAR; i.e., if a missing value is present, it seems plausible that it will be related to other patient information. Because deleting rows with missing data is not a good option for this study, we performed data imputation using the \texttt{missForest} R package (\cite{bib21}, \cite{bib22})), which is suitable for both types of missingness, MCAR and MAR. This algorithm uses a Random Forest to impute missing values and does not require any parameters to be specified.

A value zero is assigned to a diagnosis or procedure not applicable in a particular case. In our data, we are unable to distinguish whether a missing value represents such a case or whether it has not been collected. Following clinical advice, we have assumed that missing values for diagnosis and procedures represent non-applicability and have assigned a value of zero to missing values for these variables. Where the data records included wrongly inputted values in the variables describing the \emph{Age} and \emph{ethnicity} of a person, the following actions were taken:

\begin{itemize}
    \item For \emph{Age}, this was handled by converting the values over 120 to NA (indication of missing data in R) and then used the chosen imputation method. This is because clinical expertise tells us 120 normally indicates missing information on \emph{Age}. Therefore, although not the actual Age of a patient, the value contains valuable information to retain.
    \item For \emph{ethnicity}, all invalidly inputted values followed the same format and were handled consistently. for example, if category “A*” was inputted, we assume that the correct \emph{ethnicity} category is “A.”.
\end{itemize} 

\section{Learning causal models}\label{sec4}

The approach taken to construct a CBN is as follows:

\begin{enumerate}
    \item Construct a knowledge-based structure of the causal relationships between the variables present in the data using clinical expertise and existing literature.
    \item Use two different structure learning algorithms from each class of algorithms to learn the structure of BNs from data alone.
    \item Perform model-averaging on the learnt structures to obtain a model that reflects all algorithms considered.
    \item Compare the structures generated by the algorithms to the knowledge-based structure.
    \item Incorporate knowledge-based constraints into the algorithms and repeat steps 2 to 4 to obtain structures based on knowledge and data.
    \item Use the model-averaged structure with knowledge-based constraints to parametrise a CBN.
    \item Use the CBN for prediction, causal inference, and simulation of interventions.
\end{enumerate}

\subsection{Learning the structure of Bayesian Networks}

One of the objectives of this study is to investigate possible causes of \emph{Sepsis}. Structure learning algorithms can be used for this purpose. However, these algorithms cannot generally discover error-free causal structures, especially when applied to real-world data, which tend to violate some of the assumptions the algorithms make about the input data \cite{bib23}. Moreover, the following assumptions are required to interpret learnt structures as causal structures that can be converted into CBNs: 

\begin{itemize}
    \item Causal Markov condition: nodes are independent of every other node, except their descendants, given their parents. This means that knowing the direct causes of a variable of interest renders any information about its indirect causes or ancestors irrelevant when predicting that variable.
    \item Causal faithfulness: there are no independencies in the underlying probability distribution that are not those implied by the DAG. In other words, faithfulness assumes that the statistical dependencies observed in the data align with those dictated by the DAG. When assuming faithfulness, if two variables are independent in the data, this independency should be explainable by the DAG; i.e., the data should not contain any additional independencies beyond those specified by the DAG.
    \item Causal sufficiency: the observed variables include all the common causes of the variables in the data. More specifically, if two variables have a hidden common cause, then sufficiency is violated.
\end{itemize}

The above assumptions are often difficult to test in practice and are unlikely to hold for most real-world datasets, including the data used in this study. However, it is necessary to assume these conditions to perform causal inference.  

This study will use structure learning algorithms from three different classes of learning. These are:
\begin{itemize}
    \item Constraint-based: algorithms that rely on statistical tests of conditional independence to evaluate the hypothesis that variables are independent of each other, and therefore, no edge must exist between them. They also attempt to orient edges between dependent variables. 
    \item Score-based: algorithms that rely on search algorithms that explore the space of candidate graphs and objective functions that score each graph visited. They return the graph that maximises a selected objective function.
    \item Hybrid: Represented by any algorithm combining the two learning strategies above.
\end{itemize}
\hfill \break
For a comprehensive review of structure learning algorithms, please refer to \cite{bib24}, \cite{bib25} and \cite{bib26}. The algorithms used in this study are described below by learning class. In the past, constraint-based algorithms were considered more appropriate than score-based algorithms for discovering causal structures. However, recent empirical studies have shown that score-based and hybrid algorithms are often better than constraint-based algorithms for this task (\cite{bib23}, \cite{bib27}). However, it remains unclear under what circumstances one class of structure learning algorithms might be more appropriate than another, which is why we investigate all three types of algorithms in this study.
\newline

\subsubsection{Constraint-based learning}
\hfill \break

We use the well-established PC-Stable algorithm (\cite{bib28}), which is an extension of the PC algorithm (\cite{bib29}) that relaxes the PC's output sensitivity to the order of the variables as read from data. It starts from a fully connected undirected graph and checks whether each pair of variables is independent or conditionally independent. When variables are independent or conditionally independent, the edge connecting them is removed. When these tests are finished, the algorithm attempts to orient the edges that remain in the graph. Ultimately, the algorithm returns a graphical structure containing both directed and undirected edges, reflecting the theoretical limitations of orientating all edges from observational data alone.

We also use the Interleaved Incremental Association (Inter-IAMB) algorithm (\cite{bib30}), which starts by learning the Markov blankets of each node. These are used to identify the neighbours (parents and children) of each node and the parents of the children. The Markov blanket represents the nodes that make the given node conditionally independent of all others. The difference between Inter-IAMB and other flavours of IAMB is the use of forward-stepwise selection, which helps to improve the accuracy of the Markov blanket candidate set for each node. Inter-IAMB will also return some undirected edges.
\hfill \break
\subsubsection{Score-based learning}
\hfill \break

We use the classic Hill-Climbing (HC) (\cite{bib31}) algorithm, which represents the simplest type of search. The algorithm begins with an empty graph and proceeds to add, remove or re-direct edges to maximise a given objective function. When the score no longer improves, the search stops and returns the highest-scoring graph. The primary weakness of HC is that it gets stuck at the first local maxima solution it visits.

The second score-based algorithm we use is the Tabu search (\cite{bib31}), which can be viewed as an extension of HC. As with HC, Tabu often starts from an empty graph and then proceeds to add, remove or re-direct edges as in HC. The difference from HC is that Tabu will sometimes make changes that reduce the objective score in an attempt to escape a local maximum solution. While this modification often helps Tabu terminate at a graph with a higher score than the graph returned by HC, there is no guarantee that Tabu will find the optimal solution, i.e., the highest-scoring graph available in the entire search space of candidate graphs. 
\newline
\subsubsection{Hybrid algorithms}
\hfill \break

Perhaps the most popular hybrid learning algorithm is the Max-Min Hill-Climbing (MMHC) by \cite{bib32}. The learning process of MMHC can be divided into two steps. The algorithm starts by constraining the set of parents for each node, thereby restricting the number of possible graphs to be explored, and then applies HC to find the optimal structure in this reduced search space.

In addition to MMHC, we test the H2PC algorithm by \cite{bib33}. H2PC first constructs the skeleton of the graph, focusing on avoiding false missing edges to learn the local structure around variables. It then performs HC to find the optimal structure.

\subsection{Model averaging}

Recall that we employ six different structure learning algorithms spanning three distinct learning classes. Current literature has demonstrated that no algorithm is consistently superior to others under different settings (\cite{bib34}). Learnt structures are highly sensitive to the algorithm selection; this level of sensitivity means that the learnt graphs could be highly inconsistent across the different algorithms. We, therefore, also apply model averaging on the learnt graphs to obtain an overall learnt structure, in addition to the six individual graphs produced by these algorithms.

Model averaging typically involves obtaining some weighted average across a set of outputs. The process employed for model averaging in this study can be described as follows:
\begin{enumerate}
    \item Learn the structures: Generate $K$ different structures. In our study, $K=6$.
    \item Initialise edge list: Collect all edges that appear in the $K$ structures and compile them into a list.
    \item Count edge frequencies: Record the number of times, $J$, an edge appears across all structures.
    \item Sort edges by frequency: Rank all edges in descending order based on how frequently they appear across the six graphs.
    \item Add edges iteratively: Start from the most frequent edges and proceed down the ranking, and add edges to the overall graph.
    \item Handle edge orientations: If an edge has two possible orientations with equal rank, use domain knowledge to determine its orientation.
    \item Avoid cycles: If adding an edge creates a cycle, reverse the edge and attempt to add it again. If the reversed edge also creates a cycle, remove the edge entirely.
    \item Using the edge counts $J$, we create new sets of structures $L$, where each $L$ contains all edges that appear at least $J$ times; e.g., $L=2$ represents the network that includes all edges with $J\geq2$.
    \item Determine optimal averaging: Get the BIC (defined in section 4.4) score for each structure in $L$. In this study, the optimal averaged structure was $L=2$.
    
\end{enumerate}

\subsection{Knowledge-based structure and constraints}

One of the objectives of this study is to compare the learnt graphs to a knowledge-based reference graph. Figure 2 presents the knowledge graph constructed based on literature review and clinical expertise, as discussed in Section 2. Please note that, in the knowledge graph, \emph{ethnicity} remains unconnected as no evidence was found in the literature linking\emph{ethnicity} to \emph{Sepsis}. Nonetheless, we include this variable in the input data to investigate whether the graphs generated by the structure learning algorithms would be consistent with this prior knowledge. In Figure 2, this structure is divided into subgraphs to facilitate analysis. The addition of aggregator nodes facilitates visualisation by preventing edges from becoming tangled.

\begin{sidewaysfigure}
    \centering
    \includegraphics[width=\textwidth,height =\textheight, keepaspectratio]{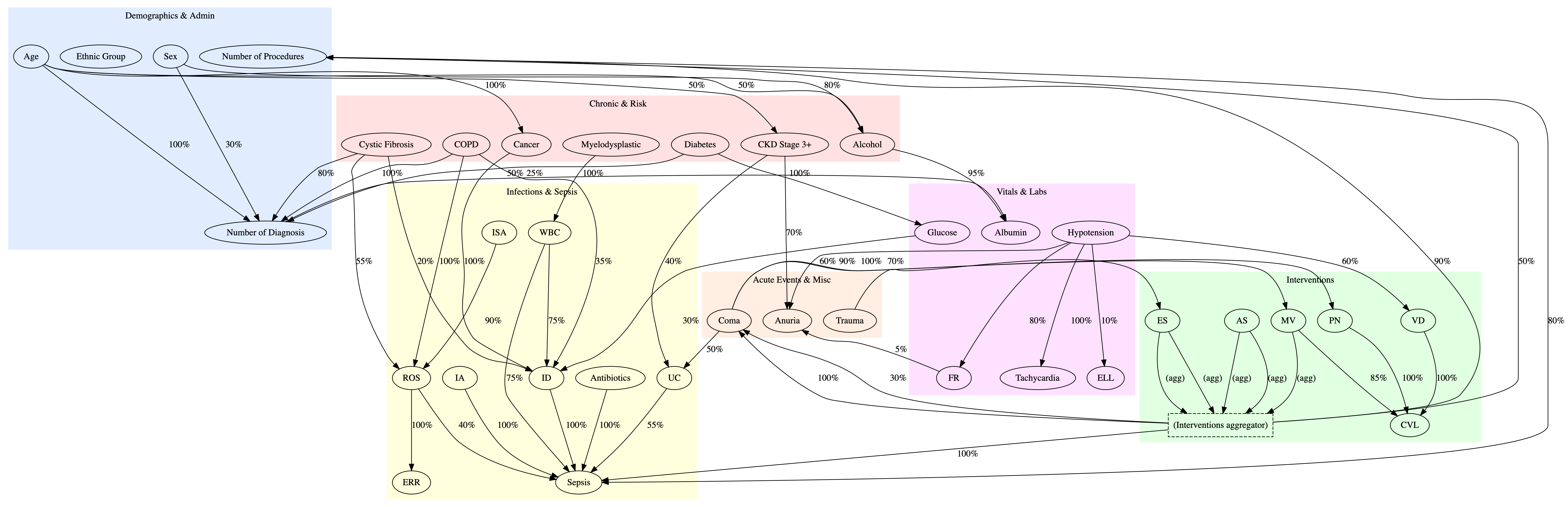}
    \caption{The knowledge DAG constructed based on literature review and clinical expertise separated with subgraphs.}
    \label{fig:enter-label}
\end{sidewaysfigure}

In addition to constructing the knowledge graph, we also produce a list of knowledge constraints consistent with it that could be incorporated into the algorithms' learning process. Table 3 presents a set of directed edges that will serve as knowledge-based constraints. We decided on these constraints by asking our clinical expert (refer to Section 2) about their confidence in the relationships of these variables. Where the answer was 90\% or higher, the constraint was generated.

The outcomes of the literature reviewed were combined with expert knowledge to construct the knowledge-based graph and assign a degree of confidence to the relationships established. The confidence assigned to the edges can be found in Figure 2. In this study, as per Figure 1, we will also assume a directed edge from \emph{Infectious Agents} to \emph{Sepsis} in the knowledge-based graph, with no directed edges being emitted by \emph{Sepsis}.

\begin{table}[h]
\caption{Directed edge constraints elicited from clinical expertise.}
\begin{tabular}{@{}ll@{}}
\toprule
\textbf{Parent node}&{\textbf{Child node}} \\
\midrule
Impaired swallow and aspiration	& Reduced oxygen saturation \\
Reduced oxygen saturation &	Elevated respiratory rate \\
Chronic Obstructive Pulmonary Disease &	Reduced oxygen saturation \\
Myelodysplastic disorders &	White blood cell counts \\
Immunosuppressive disorders	& Sepsis \\
Cancer	& Immunosuppressive disorders \\
Diabetes &	Glucose \\
Chronic Obstructive Pulmonary Disease &	Number of diagnoses \\
Alcohol dependence & Low serum albumin level\\ & at admission \\
Age &	Number of diagnoses \\
Age &	Cancer \\
Hypotension & Tachycardia \\
Vasoactive drugs &	Central venous lines \\
Parenteral nutrition & Central venous lines \\
Coma & Parenteral nutrition \\
Coma & Mechanical ventilation \\
Emergency surgeries & Number of procedures \\
Emergency surgeries & Coma \\
Mechanical ventilation & Sepsis \\
Antibiotics & Sepsis \\
Central venous lines & Sepsis\\
\botrule
\end{tabular}
\end{table}
\noindent

In addition to the directed constraints, we explore possible temporal constraints, using the principle that later events cannot cause earlier events. These temporal constraints have been used in the past by, for example, \cite{bib35}. Table 4 will define these restrictions for the learning algorithm through multiple prohibitions on directed edges. Temporal tiers are introduced to symbolise this. For example, no variables in Tiers greater than Tier 1 can cause variables in Tier 1. In addition, Tier 1 variables cannot be a cause of each other either. This means edges were forbidden within Tier 1; that is, between variables in Tier 1. However, variables in other tiers can be caused by each other; that is, edges between variables within other tiers were allowed.

\begin{table}[h]
\caption{Temporal constraints elicited from clinical expertise.}
\begin{tabular}{@{}lll@{}}
\toprule
\textbf{Tier}&{\textbf{Variables}}&{\textbf{Explanation}} \\
\midrule
1 & Age, Gender, Ethnic Group. & A person’s age, gender or ethnicity \\ & &  at admission have no cause. \\
2 & All other variables.	& 
\\
\botrule
\end{tabular}
\end{table}
\noindent

\subsection{Evaluation}
The output of each algorithm is studied in six different ways. These are:

\begin{itemize}
    \item Investigating the relationship between the learnt graphs and the reference knowledge-based graph. This is achieved using the SHD score, which counts the number of differences, also called the Hamming distance, between two graphs (\cite{bib32}). In the software used in this study, which we discuss in Section 5, the SHD metric is computed by comparing differences between Completed Partially Directed Acyclic Graphs (CPDAGs). A CPDAG contains both directed and undirected edges that cannot be orientated given the observational data and represent a set of DAGs that belong to the same Markov equivalence class.
    \item Counting the number of independent graphical fragments, or disjoint subgraphs, produced by each algorithm. This is important because information flow is not possible between independent graphical fragments, which is undesirable since we consider the input data to consist of related variables (except ethnicity).
    \item Graph complexity is determined by the number of free parameters, which represents the number of additional parameters generated by each additional edge added to the graph.
    \item The number of edges which is used to measure how dense the learnt graphs are compared to the knowledge graph.
    \item The model selection score, BIC. We use the version by \cite{bib36}, defined as:
    \begin{equation}
        BIC(G,D)= \sum_{i=1}^{p} [\log Pr (X_{i}|\prod_{X_i})-\frac{|\Theta_{X_i}|}{2}\log n]
    \end{equation}
    where $G$ denotes the graph, $D$ the data, \(X_{i}\) the nodes, \(\prod_{X_i}\)
    the parents of node \(X_{i}\), \(|\Theta_{X_i}|\) the number of free parameters in the conditional probability table, and $n$ the sample size. A higher BIC score represents a better score.
    \item The log-likelihood (LL) is a measure of how well the generated graphs fit the data.
\end{itemize}

\section{Results}
Structure learning was done using the algorithms specified in Section 4, implemented in the R package \texttt{bnlearn} (\cite{bib36}; \cite{bib37}). All of the algorithms used have hyperparameters that impact the output they create. As with previous relevant studies, we have employed the algorithms using their default hyperparameter settings, as it remains unclear in the literature how to best tune the hyperparameters in different settings or how to systematically compare results across different types of algorithms with different kinds of hyperparameters. Table 5 below shows the default \texttt{bnlearn} parameters:

\begin{table}[ht]
\caption{Default parameters for selected structure-learning algorithms in \texttt{bnlearn}.
The actual test (e.g., \texttt{mi} vs.\ \texttt{cor}) is chosen based on data type (discrete vs. continuous), where \texttt{mi} is mutual information, \texttt{restart} is the number of random restarts, setting \texttt{strict} to \texttt{TRUE} leads to a sparser network by enforcing weaker associations, and \texttt{max} hyperparameters set to \texttt{inf} indicate no constraints in the number of conditioning sets or the number of iterations in neighbouring graphs visits in constraint-based and score-based learning respectively.}
\centering
\small 
\begin{tabular}{p{3.5cm} p{2.7cm} p{7cm}}
\hline
\textbf{Algorithm \& Function} & \textbf{Default Score/Test} & \textbf{Key Default Parameters} \\
\hline
\multicolumn{3}{l}{\textbf{Score-Based}} \\
\hline
\textbf{Tabu} (\texttt{tabu()})     
  & BIC 
  & \texttt{tabu = 10}, \texttt{max.iter = Inf}, \texttt{optimized = TRUE} \\

\textbf{Hill-Climbing (HC)} (\texttt{hc()}) 
  & BIC 
  & \texttt{restart = 0}, \texttt{max.iter = Inf}, \texttt{optimized = TRUE} \\
\hline
\multicolumn{3}{l}{\textbf{Constraint-Based}} \\
\hline
\textbf{PC-stable} (\texttt{pc.stable()}) 
  & \texttt{mi} (discrete) or \texttt{cor} (continuous) 
  & \texttt{alpha = 0.05}, \texttt{strict = TRUE}, \texttt{max.sx = Inf} \\

\textbf{Inter-IAMB} (\texttt{inter.iamb()}) 
  & \texttt{mi} (discrete) or \texttt{cor} (continuous)
  & \texttt{alpha = 0.05}, \texttt{strict = TRUE}, \texttt{max.sx = Inf} \\
\hline
\multicolumn{3}{l}{\textbf{Hybrid (Constraint + Score)}} \\
\hline
\textbf{MMHC} (\texttt{mmhc()}) 
  & Restrict: \texttt{mi}, \(\alpha = 0.05\); Maximize: BIC
  & \texttt{restrict = "mmpc"}, \texttt{maximize = "hc"}, 
    \texttt{restart = 0}, \texttt{max.iter = Inf}, \texttt{optimized = TRUE} \\

\textbf{H2PC} (\texttt{h2pc()}) 
  & \texttt{mi} (discrete) or \texttt{cor} (continuous)
  & \texttt{alpha = 0.05}, \texttt{strict = TRUE}, \texttt{max.sx = Inf}, 
    \texttt{restart = 0}, \texttt{max.iter = Inf}, \texttt{optimized = TRUE} \\
\hline
\end{tabular}
\label{tab:bnlearn-defaults}
\end{table}

\subsection{Structure learning performance}

Table 6 summarises the evaluation metrics for each structure learning algorithm, the average structure obtained through the model averaging process, and the knowledge-based structure.

The results show that the constraint-based methods and MMHC did not produce a single graphical fragment, unlike the score-based algorithms  HC and Tabu, the hybrid H2PC, and the model averaging graph, which produced a single graphical fragment that connects all risk factors, which is partly explained by the high number of edges these graphs contain. It is important to remember that in the knowledge-based graph, \emph{ethnicity} is not connected to the rest of the variables as no evidence was found in the literature to consider it a risk factor of \emph{Sepsis}.

\begin{table}[h]
\caption{Summary of Results}
\begin{tabular}{@{}lllllll@{}}
\toprule
\ & &\textbf{Independent Graphical}&\textbf{Number of}&\textbf{Number of}& & \\
\textbf{Algorithms}&{\textbf{SHD}}&{\textbf{Fragments}}&{\textbf{Free Parameters}}&{\textbf{Edges}}&{\textbf{BIC}} &{\textbf{LL}} \\
\midrule
PC stable &	66	& 2	& 712 & 32 & -8540k & -8535k\\
Inter-IAMB & 66	& 3	& 4940 & 34 & -8509k & -8475k\\
HC & 105 & 1 & 3004 & 80 & -8355k & -8334k \\
TABU & 103 & 1 & 2967 & 80 & -8353k & -8332k\\
MMHC & 65 & 4 & 567 & 26 & -8518k & -8514k\\
H2PC & 99 & 1 & 2888 & 74 & -8364k & -8344k\\
Average structure & 105 & 1 & 4299 & 81 & -8366k & -8337k\\
\hline
Knowledge-based & 0 & 1 & 1146 & 50 & -8673k & -8665k\\
\botrule
\end{tabular}
\end{table}
\noindent

Regarding the number of free parameters and edges, we can see that the knowledge-based graph only generates a higher number of free parameters than the structures generated by MMHC and PC-stable. This is not a surprise given that both graphs have few edges; generally, less dense graphs generate fewer free parameters. However, this is only a general principle, as we can see how complex the structure generated by Inter-IAMB is with a small number of edges. To be more specific, there is an association between the number of free parameters and the number of edges. This is expected since more edges tend to lead to a greater number of parents per node, which can dramatically affect the number of parameters. Differences across the algorithms might be explained by a small number of variables containing many categories and how each algorithm incorporates these variables in the structures they generate.

The knowledge-based graph yields the worst result with respect to the BIC score. This means that the knowledge-based graph might not be complex enough to capture the interrelationships of variables in the data, and some important dependencies might have been missed. The fact that the knowledge graph produces a lower objective score, however, is not surprising. This is because the algorithms are designed to maximise BIC (or some other objective score), and we would generally not expect a manually constructed knowledge graph to have a better score than the highest-scoring graph an algorithm discovers.

The high SHD value produced for all algorithms implies that there are strong disagreements between the graphs produced by the algorithms and the knowledge graph produced from the literature review and clinical expertise. The lowest SHD scores come from algorithms that had the smallest number of edges. The average structure has the highest SHD value. This is expected since the averaging process will tend to include most of the edges generated from any of the algorithms, and these were already generating high SHD scores.

Overall, there seem to be two groups of algorithms with respect to the number of edges discovered. The first group consists of the two score-based learning algorithms, the hybrid H2PC, and the model averaging structure. This group has a high number of edges. On the other hand, the two constraint-based algorithms and hybrid MMHC discovered a relatively low number of edges. Most of the variables used in this study refer to the presence or absence of medical diagnosis or procedures; these are binary variables. Interestingly, this type of data structure paired with large samples, such as the one in this study, seems to generate a large number of edges in score-based methods.

BIC scores, representing a score that balances model fitting with model dimensionality, follow a similar pattern. Constraint-based algorithms are not designed to maximise BIC scores, so it is natural to expect score-based algorithms (or hybrid algorithms that employ score-based solutions) to produce better BIC scores, as evidenced in the results. Although there are some differences when considering the LL instead of the BIC score, these differences are not substantial.

Across all the six algorithms, just three edges or around 1\% of all the edges learnt by the algorithms, are discovered by all the six algorithms. These are:

\begin{enumerate}
    \item Age $\rightarrow$ Chronic Obstructive Pulmonary Disease,
    \item Number of Interventions, Operations, and Procedures $\rightarrow$ Central venous lines,
    \item Diabetes $\rightarrow$ Number of Diagnoses.
\end{enumerate}

Interestingly, only \textit{Diabetes} $\rightarrow$ \textit{Number of Diagnoses} appears in the knowledge-based graph. While this highlights considerable disagreements between knowledge and structures learnt from data, this does not necessarily imply that the different graphs are not reasonably accurate. For example, the 1st and 2nd edges from the list above present an interesting finding: these edges are deemed clinically sensible. Still, our clinical expert would not have placed a higher than 90\% confidence that these were causal relationships. This highlights the need to consider both expertise and data-driven approaches to derive new understanding.

Furthermore, there are eight edges, representing around 3\% of all the edges learnt by the algorithms, that are learnt by 5 out of the six algorithms. Two of these edges appear in the knowledge-based graph, i.e., \textit{Age} $\rightarrow$ \textit{Alcohol dependence} and \textit{Age} $\rightarrow$ \textit{Number of Diagnosis}. The algorithms learnt \textit{Alcohol dependence} $\rightarrow$ \textit{Sex} instead of \textit{Sex} $\rightarrow$ \textit{Alcohol dependence}.

Figure 3 presents a heatmap highlighting the edges that appear two or more times in the graphs learnt by the six different structure learning algorithms. We observe that the following variables had the highest numbers of direct effects (child nodes):

\begin{itemize}
    \item Number of diagnosis,
    \item Number of Interventions, Operations, and Procedures,
    \item Age,
    \item Diabetes,
    \item Cancer.
\end{itemize}

It is reasonable to expect the first two variables to emit a high number of edges since these two variables represent a collection of factors rather than a single outcome. \textit{Age} is also an important variable since as people grow older, they are expected to develop a high number of diagnoses or to have a high number of procedures. \textit{Diabetes} and \textit{Cancer} are interesting results and would mark the expected impact of these diseases on an individual's immunity. The increased frailty of an individual associated with these diagnoses can lead them to develop other conditions and diagnoses. 

The distribution for direct causes (parent nodes), shown in Figure 4, seems less skewed than the distribution for direct effects (child nodes), and no variables had large numbers of direct causes.

\begin{figure}[h]
\centerline{\includegraphics[width=\textwidth, keepaspectratio = true]{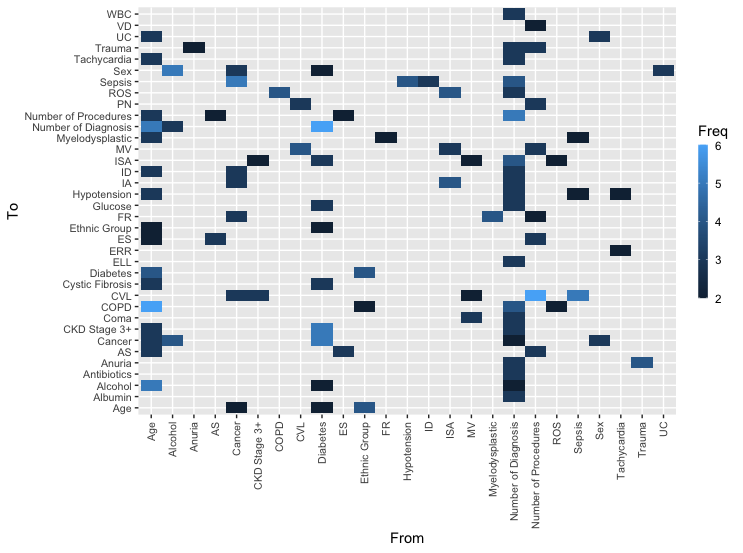}}
\caption{Heatmap of the edges that appear two or more times in the graphs learnt by the six algorithms. The figure was produced using ggplot2 (\cite{bib38}).}
\label{fig}
\end{figure}

\begin{figure}[h]
\centerline{\includegraphics[width=\textwidth, keepaspectratio = true]{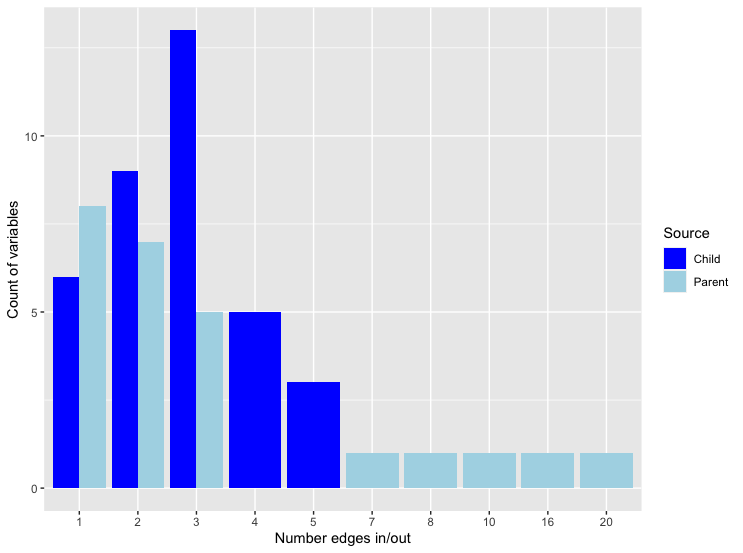}}
\caption{Distributions of the number of variables according to the number of direct effects (Number of edges in) and the number of direct causes (Number of edges out) across all the six structures learnt by the algorithms. The figure was produced using ggplot2 (\cite{bib38}).}
\label{fig}
\end{figure}

\subsection{Structure learning with knowledge-based constraints}

Table 7 summarises the evaluation metrics for each structure learning algorithm and the average structure when they incorporate the knowledge-based constraints detailed in Tables 3 and 4. Table 7 also includes information on the knowledge-based structure.

The results show that the constraints caused the SHD scores to decrease, as expected, since the constraints are guiding algorithms towards the knowledge-based graph. Moreover, the number of edges increased for all algorithms. On the other hand, the number of free parameters presents an unclear pattern since it increased for some algorithms and decreased for others. The BIC score decreased for all the structures except for the one generated by PC Stable. This might be due to increasing the complexity of a less dense structure and because constraint-based methods do not optimise BIC scores.

\begin{table}[h]
\caption{Summary of results when algorithms incorporate knowledge-based constraints}
\begin{tabular}{@{}lllllll@{}}
\toprule
\ & &\textbf{Independent Graphical}&\textbf{Number of}&\textbf{Number of}& & \\
\textbf{Algorithms}&{\textbf{SHD}}&{\textbf{Fragments}}&{\textbf{Free Parameters}}&{\textbf{Edges}}&{\textbf{BIC}}&{\textbf{LL}} \\
\midrule
PC stable &	46	& 3	& 1065 & 44 & -8528k & -8521k\\
Inter-IAMB & 51	& 3	& 2967 & 47 & -8511k & -8491k\\
HC & 81 & 1 & 2644 & 83 & -8400k & -8382k \\
TABU & 81 & 1 & 2644 & 83 & -8400k & -8382k \\
MMHC & 45 & 4 & 812 & 40 & -8546k & -8540k\\
H2PC & 76 & 1 & 2584 & 78 & -8403k & -8385k\\
Average structure & 84	& 1	& 3092 & 86 & -8398k & -8376k\\
\hline
Knowledge-based & 0 & 1 & 1146 & 50 & -8673k & -8665k\\
\botrule
\end{tabular}
\end{table}
\noindent

With the knowledge constraints enforced on all six algorithms, we obtain the model-averaging graph in Figure 5. Figure 5 shows this structure split using aggregator nodes to prevent edges from becoming tangled, facilitating analysis. We can see that the parents of \textit{Sepsis} are:

\begin{itemize}
    \item Central venous lines,
    \item Cancer,
    \item Immunosuppressive disorders,
    \item Mechanical ventilation,
    \item Number of Diagnosis,
    \item Hypotension,
    \item Antibiotics,
    \item Infectious Agents.
\end{itemize}  

Since we are applying knowledge constraints, which include constraints that assign parents to \emph{Sepsis}, it is natural that the model averaging graph with knowledge constraints will contain a higher number of parents of \emph{Sepsis} compared to the corresponding graph learnt with no constraints. An interesting finding is that \textit{Ethnic Group}, although not deemed related to \textit{Sepsis} by the literature and clinical expertise, is part of the graph and has a causal path to \textit{Sepsis}.

\begin{sidewaysfigure}
    \centering
    \includegraphics[width=\textwidth,height =\textheight, keepaspectratio]{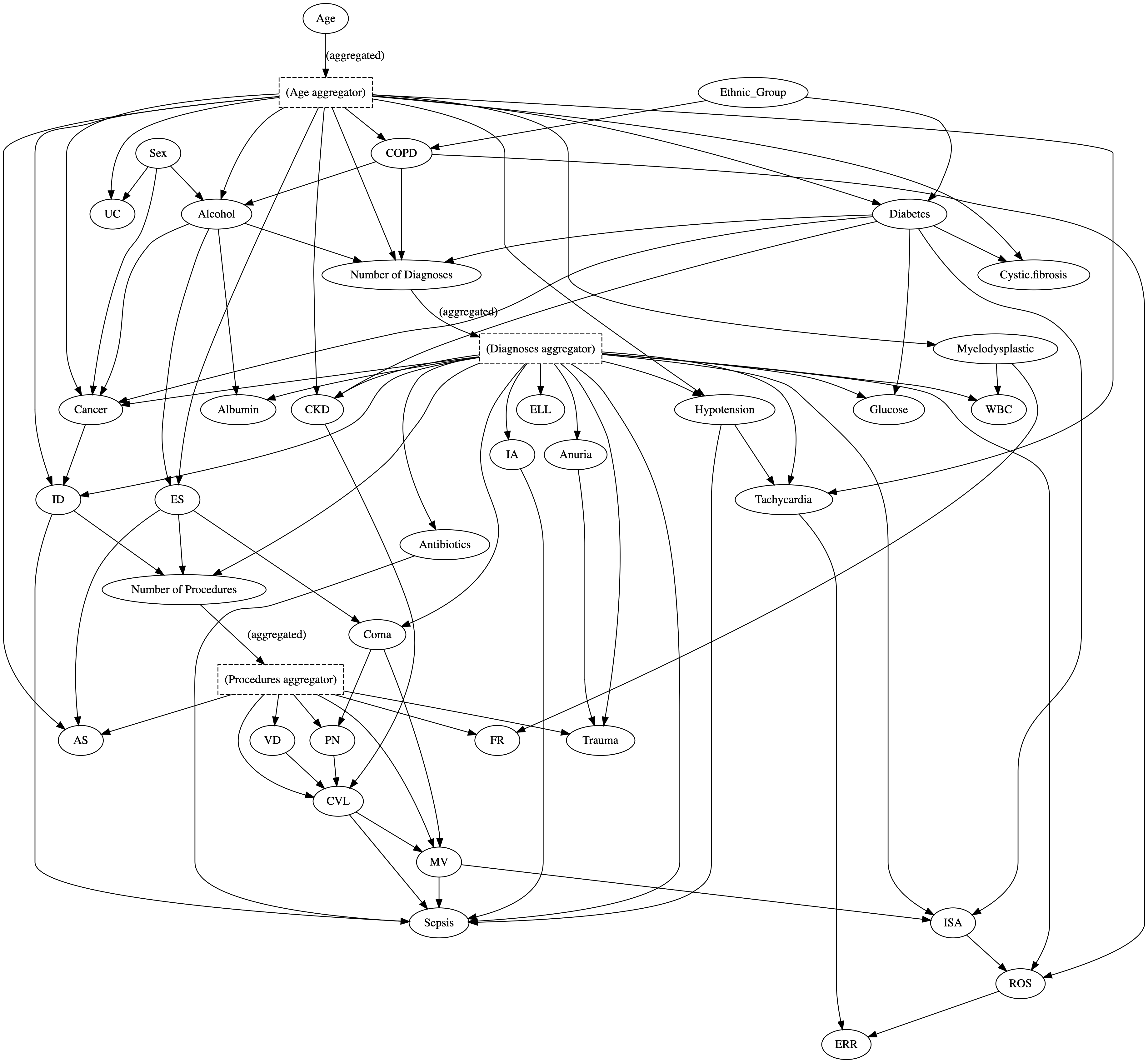}
    \caption{The structure obtained through model averaging across the six algorithms, with the knowledge constraints imposed on the structure learning process. The figure is using aggregator nodes to facilitate analysis.}
    \label{fig:enter-label}
\end{sidewaysfigure}

\subsection{Predictive validation of Sepsis}

In the field of BNs, validation methods used in structure learning usually focus on the fit of the data to the whole graph; e.g., the Log-Likelihood score. While the global probability distribution is important, we are also naturally interested in the ability of the model to predict Sepsis specifically. From the perspective of predicting \textit{Sepsis}, this structure was evaluated using the common confusion matrix, accuracy, sensitivity, and specificity indicators. The predictive validation strategy involved 10-fold cross-validation, where in every rotation, 90\% of data were used to parametrise (train) the CBN and 10\% were used for testing. The metrics presented below are an average calculated using the test data for each of the 10-fold. These metrics were used in the field of BNs by \cite{bib39} to evaluate their BN model of hypertension.

The R package \texttt{caret} (\cite{bib40}) was used to calculate the confusion matrix, accuracy, sensitivity, and specificity. The results of these indicators are calculated by judging that an episode will include a \textit{Sepsis} diagnosis if the probability of \textit{Sepsis} is higher than 3.56\% (average of \textit{Sepsis} cases in train data). This was deemed a natural threshold that could be easily adapted to other datasets. The accuracy in making predictions for our model was 69.6\%, sensitivity was 76.5\%, and specificity was 69.4\%. These results generally align with the results of the models from the literature review discussed in Section 1. The confusion matrix is shown in Table 8 below:

\begin{table}[htbp]
\caption{Confusion Matrix of the presence of Sepsis predictions on test data}
\begin{tabular}{|c|c|c|}
\hline
 & \textbf{No Sepsis (Predicted)} & \textbf{Sepsis (Predicted)} \\
\hline
\textbf{No Sepsis (Actual)} & 66.90\% & 29.53\% \\
\textbf{Sepsis (Actual)} & 0.84\% & 2.73\% \\
\hline
\end{tabular}
\label{tab1}
\end{table}

However, some of the models reviewed had a more ambitious objective: early detection of \emph{Sepsis} (\cite{bib5}, \cite{bib8}, \cite{bib9}). The input data partly explains why our model performs similarly to these models. The model presented in this paper relies on data available for commissioning purposes only. In other words, the BN models learnt in this study are trained on a subset of the data used by comparable studies. Given the data limitations and the performance metrics obtained, the predictions of the presence of \emph{Sepsis} this model can make are reasonably accurate.

Figure 6 shows a smooth and convex ROC curve arching towards the upper left corner of the plot. This is the plot we would like to see in a good predictive model since the top left corner of the plot would be the ideal model. The Area Under the Curve (AUC) is 0.8, indicating an 80\% chance that the model will be able to distinguish between \emph{Sepsis} and no \emph{Sepsis}. This is a further indication of good model performance.

\begin{figure}[htbp]
\centerline{\includegraphics[width=\textwidth, keepaspectratio = true]{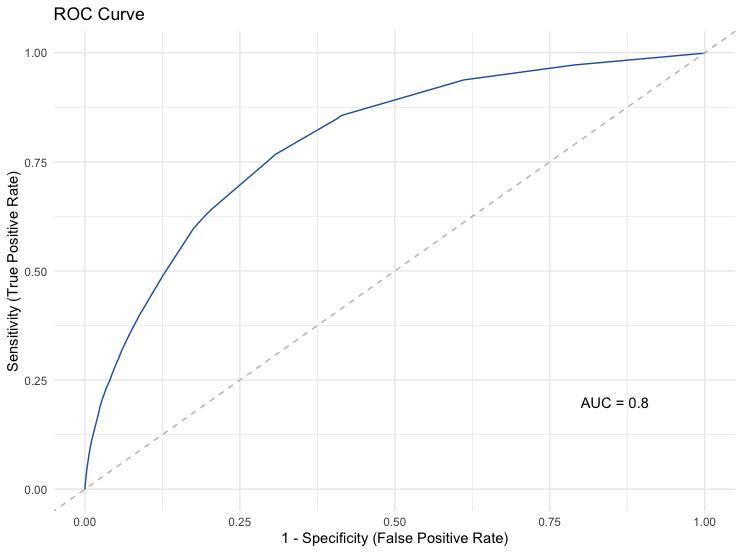}}
\caption{ROC curve of the classification problem solved with our CBN. The figure was produced using 
pROC (\cite{bib41}) and ggplot2 (\cite{bib38}).}
\label{fig}
\end{figure}

In addition to our proposed approach, several studies have developed models to predict \emph{Sepsis} or enable its early detection. These studies employed a wide range of techniques, including traditional regression models and BNs. However, all of these studies had access to broader clinical information than our dataset, which was restricted to billing data and did not include detailed clinical data such as biomarkers.

Table 9 presents the predictive performance of these models from the literature alongside our knowledge-based and averaged-structure models. Where available, we report commonly used performance metrics for classification tasks, including the Area Under the Curve (AUC), accuracy, sensitivity, and specificity. Since the studies did not explicitly name their models, we have referenced them by the first author of each work. Notably, \cite{bib8} reports multiple metrics because their model was evaluated on several datasets.

It is important to note that these comparisons are not based on identical datasets. The models differ significantly in data type, sample size, and clinical settings. Therefore, while these metrics provide insight into the relative performance of our models and those in the literature, they should be interpreted cautiously regarding direct comparisons.

The results indicate that the knowledge-based model, while designed to capture known causal relationships, performs poorly at predictive accuracy compared to the model-averaging BN model described in this study and other data-driven models in the literature. This is consistent with the finding that the knowledge graph produces the lowest objective BIC score. However, this outcome is not entirely surprising, as knowledge graphs constructed by domain experts do not maximise data fitting by design but rather aim to capture causal relationships that models maximising predictive accuracy may fail to do so. As a result, it is often the case that knowledge-based graphical structures will miss data patterns that algorithms may recover. On the other hand, the average structure, which also incorporates domain knowledge, performs comparably to other models in the literature despite being trained on a weaker dataset. Given these findings, the best approach for future work may be to combine the strengths of both methods by combining the causal insights of knowledge graphs with data-driven techniques.

\begin{table}[htbp]
\caption{\small
Comparison of predictive performance (AUC, accuracy, sensitivity, and specificity) across relevant models from the literature and those presented in this study. 
The models are listed by the first author's name as no specific model names were provided in the source papers. 
Entries left blank indicate metrics not reported in that study. 
"KB Model" refers to the knowledge-based model, and "Avg-Structure" refers to the averaged BN model.}
\label{tab:performance-comparison}
\begin{tabular}{@{}lllccc@{}}
\toprule
\textbf{Reference} & \textbf{Model / Author} & \textbf{AUC} & \textbf{Accuracy} & \textbf{Sensitivity} & \textbf{Specificity} \\
\midrule
\cite{bib5}  & Yee et al. (2019)  & 0.81             &      --         &      --         &      --         \\
\cite{bib6}  & Gupta et al. (2020)& 0.84             &      --         &      --         &      --         \\
\cite{bib7}  & Haug \& Ferraro (2016) & 0.86             &      --         & 0.76            &      --         \\
\cite{bib8}  & Pawar et al. (2019) & 0.75, 0.76, 0.78 & 0.80, 0.89, 0.82&      --         &      --         \\
\cite{bib9}  & Yang et al. (2020) & 0.97             &      --         & 0.99            & 0.75            \\
\hline
\textbf{KB Model} & (This study) 
   & 0.58     & 0.82     & 0.31     & 0.84     \\
\textbf{Avg-Structure} & (This study) 
   & 0.80     & 0.70     & 0.77     & 0.69     \\
\botrule
\end{tabular}
\end{table}

\subsection{Causal Inference and Intervention}

Simulating interventions in CBNs involves manipulating the state of a variable in a model. In this study, this is equivalent to, for example, a policy intervention to reduce alcohol consumption. Unlike assessments of predictive accuracy, estimating the effect of interventions does not require performing the train-test data split, implying that the results discussed in this subsection are based on the entire training dataset.

\cite{bib10} proposed do-calculus as a framework in CBNs to estimate the effect of hypothetical interventions without the need, for example, to perform Randomised Control Trials (RCTs). The do-calculus is represented by a do-operator \(P(Y/do(X=x))\), which can be interpreted as the probability of observing \(Y\) if we manipulate \(X\) and set it to the state \(x\). Do-calculus is a system of rules to transform expressions that contain the do-operator into expressions that do not contain it. This is equivalent to transforming a quantity we do not observe in our data to one we do. In other words, do-calculus measures the effect of a hypothetical intervention by rendering the intervened variable independent of its causes, thereby eliminating the 'explanation' for the intervention (since it was forced), which could wrongly influence its effect, and focusing entirely on measuring the effect of the intervention independently.

As shown by \cite{bib42}, do-calculus is an impressive tool in the sense that if a causal effect is identifiable, then a series of applications of the rules of do-calculus will achieve the transformation from an expression that contains the do-operator to one that does not. If no series of applications of do-calculus can do this, then we know the causal effect cannot be identified.

The R package \texttt{causaleffect} (\cite{bib43}) allows us to identify causal effects. This package implements \cite{bib44} and \cite{bib45}. If an effect is identifiable, an expression will be returned on how to estimate the causal effect. We then implement this calculation using the functionalities of \texttt{bnlearn}.

A simple review of the variables used in this study tells us that \textit{Alcohol dependence}, \textit{Chronic Obstructive Pulmonary Disease} (\textit{COPD}), and \textit{Diabetes} are interesting variables to consider for intervention. One of the leading causes of \textit{COPD} is smoking which, together with alcohol consumption, has been widely and publicly targeted by policymakers to achieve a reduction in consumption. One of the main risk factors of type II diabetes is obesity, which might also be intervened by policy action. This study will not explore how, for example, a reduction in smoking will impact the prevalence of \textit{COPD}.

We estimated that:
\begin{itemize}
    \item Intervening on \textit{COPD} causes \textit{Sepsis} to drop from approximately 5.5\% when a patient has \textit{COPD} to approximately 3.4\% when a patient does not have \textit{COPD}, which is a decrease of 2.1\% (38.2\% relative decrease).
    \item Intervening on \textit{Alcohol dependence} causes \textit{Sepsis} to drop from approximately 5.4\% when a patient has \textit{Alcohol dependence} to approximately 3.6\% when a patient does not have \textit{Alcohol dependence}, which is a decrease of 1.8\% (33.3\% relative decrease).
    \item Intervening on \textit{Diabetes} causes \textit{Sepsis} to drop from approximately 4.9\% when a patient has \textit{Diabetes} to approximately 3.2\% when a patient does not have \textit{Diabetes}, which is a decrease of 1.7\% (34.7\% relative decrease).
\end{itemize}

These results estimate the potential benefits of preventing these factors for the occurrence of \textit{Sepsis}. Of course, there might be other benefits of this prevention not considered here. However, these results show the relevance of modelling the interrelations of these risk factors in understanding how information is transmitted across the system to enable the discovery of possible interventions to reduce \textit{Sepsis} cases. 

\subsubsection{Actionable Policy Recommendations}

In light of these findings, we highlight the following evidence-based policy recommendations that have the potential to reduce \textit{COPD}, \textit{Alcohol dependence}, and \textit{Diabetes}, thereby reducing the incidence of \textit{Sepsis}:

\begin{itemize}
     \item Investing in smoking cessation programmes. For example, providing subsidised access to nicotine replacement therapies.
     \item Screening for alcohol misuse and offering referral pathways to specialised addiction services. Increasing educational programmes to address alcohol misuse before it develops into dependence.
     \item Expanding lifestyle modification programmes addressing balanced nutrition, regular physical activity, and weight management for patients with diabetes.
     \item Consider increasing taxation on tobacco and alcohol products.
\end{itemize}

These recommendations not only aim to reduce \textit{COPD}, \textit{Alcohol dependence}, and \textit{Diabetes}, but also contribute to lowering the burden of \textit{Sepsis} and other related comorbidities. Investing in targeted interventions enables healthcare systems to better realise broader benefits for population health. Structural models such as those considered in this study enable improved modelling of the interrelations of risk factors and could identify and better estimate the impact of these interventions and their resulting broader benefits.

\section{Conclusions}

This study focused on constructing CBN models of \textit{Sepsis} using data routinely collected by the NHS in England for commissioning purposes and prior knowledge obtained from the literature and through clinical expertise. We have employed structure learning algorithms spanning different classes of learning to explore how the graphs they discover might differ, as well as how they might differ from the knowledge graph produced through clinical expertise.

The algorithms that perform structure learning can capture interrelations among variables. Since the interrelations captured are sensitive to the algorithms used, model averaging was performed to investigate how the overall graph learnt from data compares to clinical expertise. We find that the graphs learnt from data alone differ considerably from those produced through clinical expertise, but this does not necessarily imply that some of these graphs must be wrong or not useful. In other words, many of the edges produced by the algorithms are deemed clinically sensible despite not being present in the knowledge graph we initially produced. This highlights the need to consider both expertise and data-driven approaches to derive new understanding.

We also explored the effect of introducing knowledge-based constraints to the structure learning process of these algorithms. The constraints imposed were those in which our clinical expert had at least 90\% confidence that they were correct. While the constraints have naturally guided the algorithms towards the knowledge graph, they enabled us to investigate how these constrained graphs differed from the purely data-driven graphs and how they differ from the knowledge graph that contains both high and low-confidence edges. Although it is difficult to justify causality in this instance, assuming it allowed us to estimate the average causal effect of presence and non-presence of \textit{COPD}, \textit{Alcohol dependence} and \textit{Diabetes} on \textit{Sepsis}. The information obtained could help inform policy on the benefits of reducing the incidence of these risk factors in the population. Moreover, \textit{Ethnicity} was found to be in the causal pathway to \textit{Sepsis}, something we wanted to explore as it was absent from the literature. The edges \textit{Age} $\rightarrow$ \textit{Chronic Obstructive Pulmonary Disease} and \textit{Number of Interventions, Operations, and Procedures} $\rightarrow$ \textit{Central venous lines} were consistently discovered by the algorithms. However, clinical expertise deemed these edges reasonable, but the confidence in their existence was not as high as for other edges.

This study comes with some limitations that guide future research directions. Firstly, the data available for this analysis is a snapshot of hospital episodes. The clinical expert stated that it is possible that a patient with a history of \emph{Sepsis} might be at a higher risk of re-occurrence of \emph{Sepsis} and quantified this belief with 40\% confidence. Furthermore, the data used lacks relevant clinical features such as biomarkers, lab results or medication dosages, which could improve model performance. A richer dataset could enable us to incorporate levels of severity, past diagnosis information, or build time-varying features such as changes in lactate levels over time.

Secondly, it is important to reiterate that the results presented in this study assume that the edges produced by the algorithms and the clinical expert represent causal relationships, which is a necessary assumption for causal inference. Enforcing certain relationships based on clinical expertise to refine the structure learnt by the algorithms may be subject to other forms of bias, and the inclusion of a literature review aimed to mitigate this risk. Future work could explore more sophisticated ways to integrate expert knowledge, by expanding the datasets with richer clinical features, which would allow for more detailed causal representations, thereby improving the overall model performance.

Lastly, the missing data (0.6\%) was handled using the well-established missForest R package. While the small percentage of missing values and the large sample size (1 million hospital episodes) suggest minimal impact on the study's findings, alternative imputation methods were not explored. Moreover, since the data used in this study comes from hospitals in England, future research could explore whether these results are consistent with those obtained from hospitals in other countries, potentially identifying important differences between countries exercising similar or different relevant practices, and enhancing the generalisability of the model.

\section*{Acknowledgment}

The first author did the analysis while employed by NHS Midlands and Lancashire Commissioning Support Unit. Although this constituted a personal research project and the organisation was not consulted on content, we would like to thank the support of the organisation to carry out this project under the existing license to use the data set.
We thank Dr Jonathon Dean for his contributions in reviewing the data used in the model and for making recommendations on extra variables. We also thank him for his guidance in building the variables from clinical codes. We thank him for building the knowledge-based structure with us and for providing his confidence in the relationships established. We also thank him for reviewing the results to discuss whether the edges found were clinically sensible.

\section*{Declarations}

\subsection{Data availability and access}

The data used in this study is the Secondary Uses Service (SUS+) database from NHS England. This data was accessed with the purpose of doing this research via the National Commissioning Data Repository (NCDR), a pseudonymised patient-level data repository managed by NHS England. This data is not publicly available.

\subsection{Competing interests}

The authors declare that they have no known competing financial interests or personal relationships that could have appeared to influence the work reported in this paper.

\subsection{Ethical and informed consent for data used}

This article does not contain any studies with human participants or animals performed by any of the authors.

\subsection{Contributions}

Bruno Petrungaro: Conceptualization, Methodology, Software, Analysis, Writing-original draft preparation, review and editing. 
Neville K. Kitson: Software, Writing-review and editing.
Anthony C. Constantinou: Methodology, Writing - Review \& Editing, Supervision.

\bibliography{sn-article}


\begin{thebibliography}{45}
\ifx \bisbn   \undefined \def \bisbn  #1{ISBN #1}\fi
\ifx \binits  \undefined \def \binits#1{#1}\fi
\ifx \bauthor  \undefined \def \bauthor#1{#1}\fi
\ifx \batitle  \undefined \def \batitle#1{#1}\fi
\ifx \bjtitle  \undefined \def \bjtitle#1{#1}\fi
\ifx \bvolume  \undefined \def \bvolume#1{\textbf{#1}}\fi
\ifx \byear  \undefined \def \byear#1{#1}\fi
\ifx \bissue  \undefined \def \bissue#1{#1}\fi
\ifx \bfpage  \undefined \def \bfpage#1{#1}\fi
\ifx \blpage  \undefined \def \blpage #1{#1}\fi
\ifx \burl  \undefined \def \burl#1{\textsf{#1}}\fi
\ifx \doiurl  \undefined \def \doiurl#1{\url{https://doi.org/#1}}\fi
\ifx \betal  \undefined \def \betal{\textit{et al.}}\fi
\ifx \binstitute  \undefined \def \binstitute#1{#1}\fi
\ifx \binstitutionaled  \undefined \def \binstitutionaled#1{#1}\fi
\ifx \bctitle  \undefined \def \bctitle#1{#1}\fi
\ifx \beditor  \undefined \def \beditor#1{#1}\fi
\ifx \bpublisher  \undefined \def \bpublisher#1{#1}\fi
\ifx \bbtitle  \undefined \def \bbtitle#1{#1}\fi
\ifx \bedition  \undefined \def \bedition#1{#1}\fi
\ifx \bseriesno  \undefined \def \bseriesno#1{#1}\fi
\ifx \blocation  \undefined \def \blocation#1{#1}\fi
\ifx \bsertitle  \undefined \def \bsertitle#1{#1}\fi
\ifx \bsnm \undefined \def \bsnm#1{#1}\fi
\ifx \bsuffix \undefined \def \bsuffix#1{#1}\fi
\ifx \bparticle \undefined \def \bparticle#1{#1}\fi
\ifx \barticle \undefined \def \barticle#1{#1}\fi
\bibcommenthead
\ifx \bconfdate \undefined \def \bconfdate #1{#1}\fi
\ifx \botherref \undefined \def \botherref #1{#1}\fi
\ifx \url \undefined \def \url#1{\textsf{#1}}\fi
\ifx \bchapter \undefined \def \bchapter#1{#1}\fi
\ifx \bbook \undefined \def \bbook#1{#1}\fi
\ifx \bcomment \undefined \def \bcomment#1{#1}\fi
\ifx \oauthor \undefined \def \oauthor#1{#1}\fi
\ifx \citeauthoryear \undefined \def \citeauthoryear#1{#1}\fi
\ifx \endbibitem  \undefined \def \endbibitem {}\fi
\ifx \bconflocation  \undefined \def \bconflocation#1{#1}\fi
\ifx \arxivurl  \undefined \def \arxivurl#1{\textsf{#1}}\fi
\csname PreBibitemsHook\endcsname

\bibitem[\protect\citeauthoryear{Singer et~al.}{2016}]{bib1}
\begin{barticle}
\bauthor{\bsnm{Singer}, \binits{M.}},
\bauthor{\bsnm{Deutschman}, \binits{C.S.}},
\bauthor{\bsnm{Seymour}, \binits{C.W.}},
\bauthor{\bsnm{Shankar-Hari}, \binits{M.}},
\bauthor{\bsnm{Annane}, \binits{D.}},
\bauthor{\bsnm{Bauer}, \binits{M.}},
\bauthor{\bsnm{Bellomo}, \binits{R.}},
\bauthor{\bsnm{Bernard}, \binits{G.R.}},
\bauthor{\bsnm{Chiche}, \binits{J.-D.}},
\bauthor{\bsnm{Coopersmith}, \binits{C.M.}}, \betal:
\batitle{The third international consensus definitions for sepsis and septic
  shock (sepsis-3)}.
\bjtitle{Jama}
\bvolume{315}(\bissue{8}),
\bfpage{801}--\blpage{810}
(\byear{2016})
\end{barticle}
\endbibitem

\bibitem[\protect\citeauthoryear{Kumar}{2020}]{bib2}
\begin{barticle}
\bauthor{\bsnm{Kumar}, \binits{V.}}:
\batitle{Sepsis roadmap: what we know, what we learned, and where we are
  going}.
\bjtitle{Clinical Immunology}
\bvolume{210},
\bfpage{108264}
(\byear{2020})
\end{barticle}
\endbibitem

\bibitem[\protect\citeauthoryear{Daniels and Nutbeam}{2009}]{bib3}
\begin{bbook}
\bauthor{\bsnm{Daniels}, \binits{R.}},
\bauthor{\bsnm{Nutbeam}, \binits{T.}}:
\bbtitle{ABC of Sepsis}.
\bpublisher{John Wiley \& Sons},
\blocation{Chichester,UK}
(\byear{2009})
\end{bbook}
\endbibitem

\bibitem[\protect\citeauthoryear{Gultepe et~al.}{2012}]{bib4}
\begin{bchapter}
\bauthor{\bsnm{Gultepe}, \binits{E.}},
\bauthor{\bsnm{Nguyen}, \binits{H.}},
\bauthor{\bsnm{Albertson}, \binits{T.}},
\bauthor{\bsnm{Tagkopoulos}, \binits{I.}}:
\bctitle{A bayesian network for early diagnosis of sepsis patients: a basis for
  a clinical decision support system}.
In: \bbtitle{2012 IEEE 2nd International Conference on Computational Advances
  in Bio and Medical Sciences (ICCABS)},
pp. \bfpage{1}--\blpage{5}
(\byear{2012}).
\bcomment{IEEE}
\end{bchapter}
\endbibitem

\bibitem[\protect\citeauthoryear{Yee et~al.}{2019}]{bib5}
\begin{barticle}
\bauthor{\bsnm{Yee}, \binits{C.R.}},
\bauthor{\bsnm{Narain}, \binits{N.R.}},
\bauthor{\bsnm{Akmaev}, \binits{V.R.}},
\bauthor{\bsnm{Vemulapalli}, \binits{V.}}:
\batitle{A data-driven approach to predicting septic shock in the intensive
  care unit}.
\bjtitle{Biomedical informatics insights}
\bvolume{11},
\bfpage{1178222619885147}
(\byear{2019})
\end{barticle}
\endbibitem

\bibitem[\protect\citeauthoryear{Gupta et~al.}{2020}]{bib6}
\begin{barticle}
\bauthor{\bsnm{Gupta}, \binits{A.}},
\bauthor{\bsnm{Liu}, \binits{T.}},
\bauthor{\bsnm{Shepherd}, \binits{S.}}:
\batitle{Clinical decision support system to assess the risk of sepsis using
  tree augmented bayesian networks and electronic medical record data}.
\bjtitle{Health informatics journal}
\bvolume{26}(\bissue{2}),
\bfpage{841}--\blpage{861}
(\byear{2020})
\end{barticle}
\endbibitem

\bibitem[\protect\citeauthoryear{Haug and Ferraro}{2016}]{bib7}
\begin{bchapter}
\bauthor{\bsnm{Haug}, \binits{P.}},
\bauthor{\bsnm{Ferraro}, \binits{J.}}:
\bctitle{Using a semi-automated modeling environment to construct a bayesian,
  sepsis diagnostic system}.
In: \bbtitle{Proceedings of the 7th ACM International Conference on
  Bioinformatics, Computational Biology, and Health Informatics},
pp. \bfpage{571}--\blpage{578}
(\byear{2016})
\end{bchapter}
\endbibitem

\bibitem[\protect\citeauthoryear{Pawar et~al.}{2019}]{bib8}
\begin{bchapter}
\bauthor{\bsnm{Pawar}, \binits{R.}},
\bauthor{\bsnm{Bone}, \binits{J.}},
\bauthor{\bsnm{Ansermino}, \binits{J.M.}},
\bauthor{\bsnm{G{\"o}rges}, \binits{M.}}:
\bctitle{An algorithm for early detection of sepsis using traditional
  statistical regression modeling}.
In: \bbtitle{2019 Computing in Cardiology (CinC)},
p. \bfpage{1}
(\byear{2019}).
\bcomment{IEEE}
\end{bchapter}
\endbibitem

\bibitem[\protect\citeauthoryear{Yang et~al.}{2020}]{bib9}
\begin{barticle}
\bauthor{\bsnm{Yang}, \binits{J.}},
\bauthor{\bsnm{Ma}, \binits{Y.}},
\bauthor{\bsnm{Mao}, \binits{M.}},
\bauthor{\bsnm{Zhang}, \binits{P.}},
\bauthor{\bsnm{Gao}, \binits{H.}}:
\batitle{Application of regression model combined with computer technology in
  the construction of early warning model of sepsis infection in children}.
\bjtitle{Journal of infection and public health}
\bvolume{13}(\bissue{2}),
\bfpage{253}--\blpage{259}
(\byear{2020})
\end{barticle}
\endbibitem

\bibitem[\protect\citeauthoryear{Pearl}{1995}]{bib10}
\begin{barticle}
\bauthor{\bsnm{Pearl}, \binits{J.}}:
\batitle{Causal diagrams for empirical research}.
\bjtitle{Biometrika}
\bvolume{82}(\bissue{4}),
\bfpage{669}--\blpage{688}
(\byear{1995})
\end{barticle}
\endbibitem

\bibitem[\protect\citeauthoryear{Trentzsch et~al.}{2014}]{bib11}
\begin{barticle}
\bauthor{\bsnm{Trentzsch}, \binits{H.}},
\bauthor{\bsnm{Nienaber}, \binits{U.}},
\bauthor{\bsnm{Behnke}, \binits{M.}},
\bauthor{\bsnm{Lefering}, \binits{R.}},
\bauthor{\bsnm{Piltz}, \binits{S.}}:
\batitle{Female sex protects from organ failure and sepsis after major trauma
  haemorrhage}.
\bjtitle{Injury}
\bvolume{45},
\bfpage{20}--\blpage{28}
(\byear{2014})
\end{barticle}
\endbibitem

\bibitem[\protect\citeauthoryear{Bar{\v{s}}i{\'c} et~al.}{1999}]{bib12}
\begin{barticle}
\bauthor{\bsnm{Bar{\v{s}}i{\'c}}, \binits{B.}},
\bauthor{\bsnm{Beus}, \binits{I.}},
\bauthor{\bsnm{Marton}, \binits{E.}},
\bauthor{\bsnm{Himbele}, \binits{J.}},
\bauthor{\bsnm{Klinar}, \binits{I.}}:
\batitle{Nosocomial infections in critically ill infectious disease patients:
  results of a 7-year focal surveillance}.
\bjtitle{Infection}
\bvolume{27},
\bfpage{16}--\blpage{22}
(\byear{1999})
\end{barticle}
\endbibitem

\bibitem[\protect\citeauthoryear{Elias et~al.}{2012}]{bib13}
\begin{barticle}
\bauthor{\bsnm{Elias}, \binits{A.C.G.P.}},
\bauthor{\bsnm{Matsuo}, \binits{T.}},
\bauthor{\bsnm{Grion}, \binits{C.M.C.}},
\bauthor{\bsnm{Cardoso}, \binits{L.T.Q.}},
\bauthor{\bsnm{Verri}, \binits{P.H.}}:
\batitle{Incidence and risk factors for sepsis in surgical patients: a cohort
  study}.
\bjtitle{Journal of critical care}
\bvolume{27}(\bissue{2}),
\bfpage{159}--\blpage{166}
(\byear{2012})
\end{barticle}
\endbibitem

\bibitem[\protect\citeauthoryear{Farinas-Alvarez et~al.}{2000}]{bib14}
\begin{barticle}
\bauthor{\bsnm{Farinas-Alvarez}, \binits{C.}},
\bauthor{\bsnm{Farinas}, \binits{M.}},
\bauthor{\bsnm{Fernandez-Mazarrasa}, \binits{C.}},
\bauthor{\bsnm{Llorca}, \binits{J.}},
\bauthor{\bsnm{Casanova}, \binits{D.}},
\bauthor{\bsnm{Delgado-Rodr{\'\i}guez}, \binits{M.}}:
\batitle{Analysis of risk factors for nosocomial sepsis in surgical patients}.
\bjtitle{Journal of British Surgery}
\bvolume{87}(\bissue{8}),
\bfpage{1076}--\blpage{1081}
(\byear{2000})
\end{barticle}
\endbibitem

\bibitem[\protect\citeauthoryear{Berger et~al.}{2014}]{bib15}
\begin{barticle}
\bauthor{\bsnm{Berger}, \binits{B.}},
\bauthor{\bsnm{Gumbinger}, \binits{C.}},
\bauthor{\bsnm{Steiner}, \binits{T.}},
\bauthor{\bsnm{Sykora}, \binits{M.}}:
\batitle{Epidemiologic features, risk factors, and outcome of sepsis in stroke
  patients treated on a neurologic intensive care unit}.
\bjtitle{Journal of critical care}
\bvolume{29}(\bissue{2}),
\bfpage{241}--\blpage{248}
(\byear{2014})
\end{barticle}
\endbibitem

\bibitem[\protect\citeauthoryear{O’Brien~Jr et~al.}{2007}]{bib16}
\begin{barticle}
\bauthor{\bsnm{O’Brien~Jr}, \binits{J.M.}},
\bauthor{\bsnm{Lu}, \binits{B.}},
\bauthor{\bsnm{Ali}, \binits{N.A.}},
\bauthor{\bsnm{Martin}, \binits{G.S.}},
\bauthor{\bsnm{Aberegg}, \binits{S.K.}},
\bauthor{\bsnm{Marsh}, \binits{C.B.}},
\bauthor{\bsnm{Lemeshow}, \binits{S.}},
\bauthor{\bsnm{Douglas}, \binits{I.S.}}:
\batitle{Alcohol dependence is independently associated with sepsis, septic
  shock, and hospital mortality among adult intensive care unit patients}.
\bjtitle{Critical care medicine}
\bvolume{35}(\bissue{2}),
\bfpage{345}--\blpage{350}
(\byear{2007})
\end{barticle}
\endbibitem

\bibitem[\protect\citeauthoryear{Raleigh and Holmes}{2021}]{bib17}
\begin{bbook}
\bauthor{\bsnm{Raleigh}, \binits{V.S.}},
\bauthor{\bsnm{Holmes}, \binits{J.}}:
\bbtitle{The Health of People from Ethnic Minority Groups in England}.
\bpublisher{King's Fund},
\blocation{https://www.kingsfund.org.uk/insight-and-analysis/long-reads/health-people-ethnic-minority-groups-england\#conclusion
  [Accessed July 2021]}
(\byear{2021})
\end{bbook}
\endbibitem

\bibitem[\protect\citeauthoryear{Scheuner et~al.}{1997}]{bib18}
\begin{barticle}
\bauthor{\bsnm{Scheuner}, \binits{M.T.}},
\bauthor{\bsnm{Wang}, \binits{S.-J.}},
\bauthor{\bsnm{Raffel}, \binits{L.J.}},
\bauthor{\bsnm{Larabell}, \binits{S.K.}},
\bauthor{\bsnm{Rotter}, \binits{J.I.}}:
\batitle{Family history: a comprehensive genetic risk assessment method for the
  chronic conditions of adulthood}.
\bjtitle{American journal of medical genetics}
\bvolume{71}(\bissue{3}),
\bfpage{315}--\blpage{324}
(\byear{1997})
\end{barticle}
\endbibitem

\bibitem[\protect\citeauthoryear{{World Health Organization}}{2016}]{bib19}
\begin{botherref}
\oauthor{\bsnm{{World Health Organization}}}:
International Statistical Classification of Diseases and Related Health
  Problems,
10th revision, 2016 edition edn.
(2016)
\end{botherref}
\endbibitem

\bibitem[\protect\citeauthoryear{Hester and Wickham}{2021}]{bib20}
\begin{botherref}
\oauthor{\bsnm{Hester}, \binits{J.}},
\oauthor{\bsnm{Wickham}, \binits{H.}}:
Connect to odbc compatible databases (using the dbi interface).
Available at: https://cran.r-project.org/web/packages/odbc/ [Accessed July
  2021]
(2021)
\end{botherref}
\endbibitem

\bibitem[\protect\citeauthoryear{Stekhoven and B{\"u}hlmann}{2012}]{bib21}
\begin{barticle}
\bauthor{\bsnm{Stekhoven}, \binits{D.J.}},
\bauthor{\bsnm{B{\"u}hlmann}, \binits{P.}}:
\batitle{Missforest—non-parametric missing value imputation for mixed-type
  data}.
\bjtitle{Bioinformatics}
\bvolume{28}(\bissue{1}),
\bfpage{112}--\blpage{118}
(\byear{2012})
\end{barticle}
\endbibitem

\bibitem[\protect\citeauthoryear{Stekhoven}{2016}]{bib22}
\begin{botherref}
\oauthor{\bsnm{Stekhoven}, \binits{D.J.}}:
missForest: Nonparametric Missing Value Imputation Using Random Forest.
(2016).
Available at: https://cran.r-project.org/web/packages/missForest/index.html
\end{botherref}
\endbibitem

\bibitem[\protect\citeauthoryear{Constantinou et~al.}{2021}]{bib23}
\begin{barticle}
\bauthor{\bsnm{Constantinou}, \binits{A.C.}},
\bauthor{\bsnm{Liu}, \binits{Y.}},
\bauthor{\bsnm{Chobtham}, \binits{K.}},
\bauthor{\bsnm{Guo}, \binits{Z.}},
\bauthor{\bsnm{Kitson}, \binits{N.K.}}:
\batitle{Large-scale empirical validation of bayesian network structure
  learning algorithms with noisy data}.
\bjtitle{International Journal of Approximate Reasoning}
\bvolume{131},
\bfpage{151}--\blpage{188}
(\byear{2021})
\end{barticle}
\endbibitem

\bibitem[\protect\citeauthoryear{Kitson et~al.}{2023}]{bib24}
\begin{barticle}
\bauthor{\bsnm{Kitson}, \binits{N.K.}},
\bauthor{\bsnm{Constantinou}, \binits{A.C.}},
\bauthor{\bsnm{Guo}, \binits{Z.}},
\bauthor{\bsnm{Liu}, \binits{Y.}},
\bauthor{\bsnm{Chobtham}, \binits{K.}}:
\batitle{A survey of bayesian network structure learning}.
\bjtitle{Artificial Intelligence Review}
\bvolume{56}(\bissue{8}),
\bfpage{8721}--\blpage{8814}
(\byear{2023})
\end{barticle}
\endbibitem

\bibitem[\protect\citeauthoryear{Squires and Uhler}{2023}]{bib25}
\begin{barticle}
\bauthor{\bsnm{Squires}, \binits{C.}},
\bauthor{\bsnm{Uhler}, \binits{C.}}:
\batitle{Causal structure learning: A combinatorial perspective}.
\bjtitle{Foundations of Computational Mathematics}
\bvolume{23}(\bissue{5}),
\bfpage{1781}--\blpage{1815}
(\byear{2023})
\end{barticle}
\endbibitem

\bibitem[\protect\citeauthoryear{Vowels et~al.}{2022}]{bib26}
\begin{barticle}
\bauthor{\bsnm{Vowels}, \binits{M.J.}},
\bauthor{\bsnm{Camgoz}, \binits{N.C.}},
\bauthor{\bsnm{Bowden}, \binits{R.}}:
\batitle{D’ya like dags? a survey on structure learning and causal
  discovery}.
\bjtitle{ACM Computing Surveys}
\bvolume{55}(\bissue{4}),
\bfpage{1}--\blpage{36}
(\byear{2022})
\end{barticle}
\endbibitem

\bibitem[\protect\citeauthoryear{Scutari et~al.}{2019}]{bib27}
\begin{barticle}
\bauthor{\bsnm{Scutari}, \binits{M.}},
\bauthor{\bsnm{Graafland}, \binits{C.E.}},
\bauthor{\bsnm{Guti{\'e}rrez}, \binits{J.M.}}:
\batitle{Who learns better bayesian network structures: Accuracy and speed of
  structure learning algorithms}.
\bjtitle{International Journal of Approximate Reasoning}
\bvolume{115},
\bfpage{235}--\blpage{253}
(\byear{2019})
\end{barticle}
\endbibitem

\bibitem[\protect\citeauthoryear{Colombo et~al.}{2014}]{bib28}
\begin{barticle}
\bauthor{\bsnm{Colombo}, \binits{D.}},
\bauthor{\bsnm{Maathuis}, \binits{M.H.}}, \betal:
\batitle{Order-independent constraint-based causal structure learning.}
\bjtitle{J. Mach. Learn. Res.}
\bvolume{15}(\bissue{1}),
\bfpage{3741}--\blpage{3782}
(\byear{2014})
\end{barticle}
\endbibitem

\bibitem[\protect\citeauthoryear{Spirtes and Glymour}{1991}]{bib29}
\begin{barticle}
\bauthor{\bsnm{Spirtes}, \binits{P.}},
\bauthor{\bsnm{Glymour}, \binits{C.}}:
\batitle{An algorithm for fast recovery of sparse causal graphs}.
\bjtitle{Social science computer review}
\bvolume{9}(\bissue{1}),
\bfpage{62}--\blpage{72}
(\byear{1991})
\end{barticle}
\endbibitem

\bibitem[\protect\citeauthoryear{Tsamardinos et~al.}{2003}]{bib30}
\begin{bchapter}
\bauthor{\bsnm{Tsamardinos}, \binits{I.}},
\bauthor{\bsnm{Aliferis}, \binits{C.F.}},
\bauthor{\bsnm{Statnikov}, \binits{A.R.}},
\bauthor{\bsnm{Statnikov}, \binits{E.}}:
\bctitle{Algorithms for large scale markov blanket discovery.}
In: \bbtitle{FLAIRS},
vol. \bseriesno{2},
pp. \bfpage{376}--\blpage{81}
(\byear{2003})
\end{bchapter}
\endbibitem

\bibitem[\protect\citeauthoryear{Bouckaert}{1995}]{bib31}
\begin{botherref}
\oauthor{\bsnm{Bouckaert}, \binits{R.R.}}:
Bayesian belief networks: from construction to inference.
PhD thesis
(1995)
\end{botherref}
\endbibitem

\bibitem[\protect\citeauthoryear{Tsamardinos et~al.}{2006}]{bib32}
\begin{barticle}
\bauthor{\bsnm{Tsamardinos}, \binits{I.}},
\bauthor{\bsnm{Brown}, \binits{L.E.}},
\bauthor{\bsnm{Aliferis}, \binits{C.F.}}:
\batitle{The max-min hill-climbing bayesian network structure learning
  algorithm}.
\bjtitle{Machine learning}
\bvolume{65},
\bfpage{31}--\blpage{78}
(\byear{2006})
\end{barticle}
\endbibitem

\bibitem[\protect\citeauthoryear{Gasse et~al.}{2014}]{bib33}
\begin{barticle}
\bauthor{\bsnm{Gasse}, \binits{M.}},
\bauthor{\bsnm{Aussem}, \binits{A.}},
\bauthor{\bsnm{Elghazel}, \binits{H.}}:
\batitle{A hybrid algorithm for bayesian network structure learning with
  application to multi-label learning}.
\bjtitle{Expert Systems with Applications}
\bvolume{41}(\bissue{15}),
\bfpage{6755}--\blpage{6772}
(\byear{2014})
\end{barticle}
\endbibitem

\bibitem[\protect\citeauthoryear{Constantinou et~al.}{2023}]{bib34}
\begin{barticle}
\bauthor{\bsnm{Constantinou}, \binits{A.}},
\bauthor{\bsnm{Kitson}, \binits{N.K.}},
\bauthor{\bsnm{Liu}, \binits{Y.}},
\bauthor{\bsnm{Chobtham}, \binits{K.}},
\bauthor{\bsnm{Amirkhizi}, \binits{A.H.}},
\bauthor{\bsnm{Nanavati}, \binits{P.A.}},
\bauthor{\bsnm{Mbuvha}, \binits{R.}},
\bauthor{\bsnm{Petrungaro}, \binits{B.}}:
\batitle{Open problems in causal structure learning: A case study of covid-19
  in the uk}.
\bjtitle{Expert Systems with Applications}
\bvolume{234},
\bfpage{121069}
(\byear{2023})
\end{barticle}
\endbibitem

\bibitem[\protect\citeauthoryear{Constantinou and Fenton}{2018}]{bib35}
\begin{barticle}
\bauthor{\bsnm{Constantinou}, \binits{A.C.}},
\bauthor{\bsnm{Fenton}, \binits{N.}}:
\batitle{Things to know about bayesian networks: Decisions under uncertainty,
  part 2}.
\bjtitle{Significance}
\bvolume{15}(\bissue{2}),
\bfpage{19}--\blpage{23}
(\byear{2018})
\end{barticle}
\endbibitem

\bibitem[\protect\citeauthoryear{Scutari and Denis}{2021}]{bib36}
\begin{bbook}
\bauthor{\bsnm{Scutari}, \binits{M.}},
\bauthor{\bsnm{Denis}, \binits{J.-B.}}:
\bbtitle{Bayesian Networks: with Examples in R}.
\bpublisher{Chapman and Hall/CRC},
\blocation{New York, USA}
(\byear{2021})
\end{bbook}
\endbibitem

\bibitem[\protect\citeauthoryear{Scutari and Ness}{2012}]{bib37}
\begin{barticle}
\bauthor{\bsnm{Scutari}, \binits{M.}},
\bauthor{\bsnm{Ness}, \binits{R.}}:
\batitle{bnlearn: Bayesian network structure learning, parameter learning and
  inference}.
\bjtitle{R package version}
\bvolume{3},
\bfpage{805}
(\byear{2012})
\end{barticle}
\endbibitem

\bibitem[\protect\citeauthoryear{Wickham}{2016}]{bib38}
\begin{bbook}
\bauthor{\bsnm{Wickham}, \binits{H.}}:
\bbtitle{Ggplot2: Elegant Graphics for Data Analysis}.
\bpublisher{Springer},
\blocation{New York, USA}
(\byear{2016}).
\burl{https://ggplot2.tidyverse.org}
\end{bbook}
\endbibitem

\bibitem[\protect\citeauthoryear{Pan et~al.}{2019}]{bib39}
\begin{barticle}
\bauthor{\bsnm{Pan}, \binits{J.}},
\bauthor{\bsnm{Rao}, \binits{H.}},
\bauthor{\bsnm{Zhang}, \binits{X.}},
\bauthor{\bsnm{Li}, \binits{W.}},
\bauthor{\bsnm{Wei}, \binits{Z.}},
\bauthor{\bsnm{Zhang}, \binits{Z.}},
\bauthor{\bsnm{Ren}, \binits{H.}},
\bauthor{\bsnm{Song}, \binits{W.}},
\bauthor{\bsnm{Hou}, \binits{Y.}},
\bauthor{\bsnm{Qiu}, \binits{L.}}:
\batitle{Application of a tabu search-based bayesian network in identifying
  factors related to hypertension}.
\bjtitle{Medicine}
\bvolume{98}(\bissue{25}),
\bfpage{16058}
(\byear{2019})
\end{barticle}
\endbibitem

\bibitem[\protect\citeauthoryear{{Kuhn} and {Max}}{2008}]{bib40}
\begin{barticle}
\bauthor{\bsnm{{Kuhn}}},
\bauthor{\bsnm{{Max}}}:
\batitle{Building predictive models in r using the caret package}.
\bjtitle{Journal of Statistical Software}
\bvolume{28}(\bissue{5}),
\bfpage{1}--\blpage{26}
(\byear{2008})
\doiurl{10.18637/jss.v028.i05}
\end{barticle}
\endbibitem

\bibitem[\protect\citeauthoryear{Robin et~al.}{2011}]{bib41}
\begin{barticle}
\bauthor{\bsnm{Robin}, \binits{X.}},
\bauthor{\bsnm{Turck}, \binits{N.}},
\bauthor{\bsnm{Hainard}, \binits{A.}},
\bauthor{\bsnm{Tiberti}, \binits{N.}},
\bauthor{\bsnm{Lisacek}, \binits{F.}},
\bauthor{\bsnm{Sanchez}, \binits{J.-C.}},
\bauthor{\bsnm{Müller}, \binits{M.}}:
\batitle{proc: an open-source package for r and s+ to analyze and compare roc
  curves}.
\bjtitle{BMC Bioinformatics}
\bvolume{12},
\bfpage{77}
(\byear{2011})
\end{barticle}
\endbibitem

\bibitem[\protect\citeauthoryear{Huang and Valtorta}{2012}]{bib42}
\begin{botherref}
\oauthor{\bsnm{Huang}, \binits{Y.}},
\oauthor{\bsnm{Valtorta}, \binits{M.}}:
Pearl's calculus of intervention is complete.
arXiv preprint arXiv:1206.6831
(2012)
\end{botherref}
\endbibitem

\bibitem[\protect\citeauthoryear{Tikka and Karvanen}{2017}]{bib43}
\begin{barticle}
\bauthor{\bsnm{Tikka}, \binits{S.}},
\bauthor{\bsnm{Karvanen}, \binits{J.}}:
\batitle{Identifying causal effects with the {R} package {causaleffect}}.
\bjtitle{Journal of Statistical Software}
\bvolume{76}(\bissue{12}),
\bfpage{1}--\blpage{30}
(\byear{2017})
\doiurl{10.18637/jss.v076.i12}
\end{barticle}
\endbibitem

\bibitem[\protect\citeauthoryear{Shpitser and Pearl}{2012}]{bib44}
\begin{botherref}
\oauthor{\bsnm{Shpitser}, \binits{I.}},
\oauthor{\bsnm{Pearl}, \binits{J.}}:
Identification of conditional interventional distributions.
arXiv preprint arXiv:1206.6876
(2012)
\end{botherref}
\endbibitem

\bibitem[\protect\citeauthoryear{Shpitser and Pearl}{2006}]{bib45}
\begin{bchapter}
\bauthor{\bsnm{Shpitser}, \binits{I.}},
\bauthor{\bsnm{Pearl}, \binits{J.}}:
\bctitle{Identification of joint interventional distributions in recursive
  semi-markovian causal models}.
In: \bbtitle{Proceedings of the National Conference on Artificial
  Intelligence},
vol. \bseriesno{21},
p. \bfpage{1219}
(\byear{2006}).
\bcomment{Menlo Park, CA; Cambridge, MA; London; AAAI Press; MIT Press; 1999}
\end{bchapter}
\endbibitem

\end{thebibliography}

\end{document}